\documentclass[runningheads]{llncs}

\usepackage{booktabs,array}
\usepackage[final]{graphicx}
\usepackage{float}
\usepackage{subfig}
\usepackage{hyperref}
\usepackage[dvipsnames]{xcolor}
\usepackage{tikz}
\usepackage{hhline}
\usepackage{multirow}
\usepackage{adjustbox}
\usepackage{csquotes}
\usepackage{geometry}
\newcolumntype{M}[1]{>{\centering\arraybackslash}m{#1}}
\begin{document}
\title{Automatic detection of lumen and media in  the IVUS images using  U-Net with VGG16 Encoder}
%
%
\author{Chirag Balakrishna\and Sarshar Dadashzadeh\and
Sara Soltaninejad}
\authorrunning{C.Balakrishna et al}
%
\institute{Department of Computing Science, University of Alberta, Canada\\ \email{\{cbalakri, dadashza, soltanin\}@ualberta.ca}}
\maketitle              
\begin{abstract}
Coronary heart disease is one of the top rank leading cause of mortality in the world which can be because of plaque burden inside the arteries. Intravascular Ultrasound (IVUS) has been recognized as powerful imaging technology which captures the real time and high resolution images of the coronary arteries and can be used for the analysis of these plaques. The IVUS segmentation involves the extraction of two arterial walls components namely, lumen and media. In this paper, we investigate the effectiveness of Convolutional Neural Networks including U-Net to segment ultrasound scans of arteries. In particular, the proposed segmentation network was built based on the the U-Net with the VGG16 encoder. Experiments were done for evaluating the proposed segmentation architecture which show promising quantitative and qualitative results.
\end{abstract}
\section{Introduction}

Intravascular Ultrasound (IVUS) is a diagnosis medical imaging technique wherein an ultrasound probe is inserted into the arteries to capture the real time high resolution cross-sectional scan of the vessel \footnote{From \url{https://en.wikipedia.org/wiki/Intravascular-ultrasound}}. Segmentation of this type of imaging helps a practitioner identify locate of not only the vessel membranes (lumen and media), but also the atherosclerotic plaques in the vessel walls. The early detection of these area can help prevent further complications which can lead to myocardial infarction and ultimately death. This study is important because myocardial infarction is one of the leading causes of deaths and \cite{4} reported that there were 15.9 million deaths world wide in $2015$. However, IVUS segmentation is one of the challenging task in medical image analysis due to high volume of this type of image and existence of different types of artifacts. 

Several methods ranging from probabilistic to deep learning have been proposed to address this problem. For instance, \cite{5}, The authors tries to segment the walls of the coronary artery using the Fast Marching Method (FMM) which incorporates two components, the texture gradient and the gray level gradient. They also make use of gamma probability density function to model the gray level distributions in the arterial walls. The presence of artefacts in the image can greatly disturb segmentation accuracy. So, one advantage of this method \cite{5} is that it claims to handle artefacts fairly well. However, the method needs a large amount of data to verify robustness. One of the best achieved results in IVUS segmentation has been reported in \cite{21}. They proposed a region selection strategy on top of a recently proposed Extremal Region of Extremum Level (EREL) \cite{22,23} detector in order to segment the lumen and media. Their deterministic method not only runs very fast, but also can segment the 20Mhz IVUS frames accurately.

Studies \cite{6}, \cite{7}, \cite{8}, \cite{9}, \cite{11}, \cite{12} focus on neural networks. According to Google Trends, the current trends which includes the last five years indicate an increase in the adoption of neural networks to perform medical image segmentation and analysis \cite{13}. This is due to the availability of appropriate hardware such as powerful NVIDIA GPUs, NVIDIA cuDNN - a GPU accelerated deep learning toolkit and high performance convolutional neural networks like NiftyNet \cite{6}, V-Net \cite{7} and U-Net \cite{14} which are built for medical image analysis. Shengran et al in \cite{11}, applied Artificial Neural Network (ANN) for border detection in IVUS images including both vessel membrane and plaque type and burden to address mentioned problem. The first phase is border smoothing using a mean filter to achieve region of interest that includes vessel and plaque inside. The second phase is to make Region Of Interest (ROI) narrow limited to vessel area using Double ANN and then smoothing borders again to pass the output of prior step to the next ANN to detect plaque border. Methodology used in \cite{11} is ANN with $4$ layers which is evaluated by $461$ IVUS high resolution (up to $113$ um) images from $4$ subjects. $80$ IVUS images were used to train the model and $381$ images were used as testing data to cross validate, resulting in mean absolute error $1.96$ STD which is fairly tolerable. Lumen cross-section area (LCSA) correlation of testing data is highly accurate with $98.42\%$ and vessel cross-section area (VCSA) correlation is $96.48\%$ and manual compare value is $98.19\%$. plaque cross-section area (PCSA) correlation of testing data is $86.48\%$ and manual compare of PCSA is $94.25\%$ that focuses on sensitivity of shape of the curve.

In this paper we make use of a variant of the fully convolutional neural network called U-Net\cite{14}. We investigate the effectiveness of U-Net to segment gray scale IVUS images for identification of lumen and media. After that, the U-Net with VGG16\cite{15} encoder is used for detection of lumen and media in IVUS scans which relies heavily on the data augmentation. The rest of this paper is organized as follow. The second section describes the data set and the proposed network for segmentation. The experimental results and discussion is presented in section 3. The conclusion is explained in section 5.
\begin{figure}
	\centering
	\includegraphics[width=0.75\linewidth]{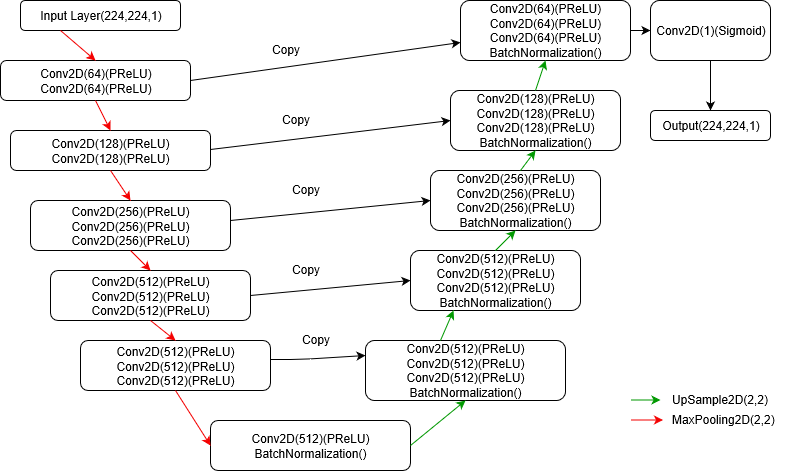}
	\caption{The VGG16-UNet architecture which is U-Net models with VGG-16 encoder.}
	\label{vggunet}
\end{figure}
\section{Method}
In this section, first the IVUS dataset that is used in this research is described. In the following, the proposed segmentation method which is based on the combination of VGG16 and U-Net architectures is discussed.  
\subsection{Dataset}
We use a publicly available dataset \cite{24} that includes ultrasound scans of the human coronary artery. 
The dimension for each scan is $384\times 384$. Dataset has $435$ scans and each has its own lumen and media labels. The training data has $109$ IVUS scans and the testing part has $326$. The test set consists of in-vivo pullbacks of the human coronary artery acquired by the $Si5$ from Volcano Corporation using the $20$MHz Eagle Eye monorail catheter (Volcano Corp., San Diego, CA). 

\subsection{Network Architecture}
The U-Net \cite{14} is one of the powerful convolutional network architecture for fast and precise images segmentation which is presented in $2015$ for the first time for biomedical image segmentation. It consists of two general parts which are encoder and decoder that makes it to have a \enquote{U} shape.  It predicts a pixel wise segmentation map of the input image rather than classifying the input image as a whole. U-Net passes the feature maps from each level of the contracting path over to the analogous level in the expanding path which allows the classifier to consider features at various scales and complexities to make its decision. U-Net is capable of learning from a relatively small training set. 


In this paper, we propose a deep learning method that is a combination of these two powerful network architectures in which the VGG-16 is used as the encoder part of the U-Net. We call the proposed architecture VGG16-UNet. The architecture is shown the Figure\ref{vggunet}. 

The left hand side of the network is an encoder and has $5$ blocks. It incorporates the $13$ convolutional layers from the original VGG16\cite{15}. After each convolution block, the red arrow indicates a MaxPooling operation which reduces the dimensions of the image by $2 \times 2$. The right hand side of the network, is a decoder which also has $5$ blocks. The green arrow after each block is an UpSampling operation which restores the dimensions of the image. Each UpSampling operation repeats the rows and columns of the image by $2 \times 2$ (two rows and columns). The skip connections between the blocks (horizontal connections with black arrows) is used to restore the dimensions of the image. These skip connections are implemented using the concatenate operation to combine the corresponding feature maps. Since this is a variant of the Fully Convolutional Neural Network, FCN\cite{16} for semantic segmentation, the spatial dimension information (height and width) of the image needs to be retained hence we use the skip connections. The last convolutional layer has only $1$ filter  which is similar to a final Dense layer in most other neural networks and gives the binary mask prediction. 
In total, the network has about $29$ convolutional layers which is followed by a PReLU\cite{17} activation. The PReLU\cite{17} has an alpha parameter that is learned during training. In addition, the last convolutional layer has a sigmoid activation function.

\subsection{Image Augmentation}
The training dataset we used comprises of $109$ IVUS scans which is not sufficient. The deep learning method's performance is depend on the amount of data. Thus, the data augmentation was performed with keeping the information in the image. The augmentations type that are performed in this paper are including:
\begin{itemize}
\item Horizontal Flips - The images were randomly flipped along the horizontal axis.
\item Vertical Flips - Images were flipped along the vertical axis.
\item Width Shift - Images were shifted along the horizontal axis by 30/width.
\item Height Shift - Images were shifted along the vertical axis by 30/height
\item Rotation - Randomly rotation of the image from 0 to 90$^{\circ}$.
\end{itemize}

\section{Experimental Results and Discussion}
Three sets of experiments were carried out for segmenting the lumen and media separately including simple U-Net model, VGG16-UNet without data augmentation (VGG16-UNet without DA), VGG16-UNet with data augmentation (VGG16-UNet with DA). In this section, first the metrics for evaluation the proposed networks are described and in the following the experiments, numerical implementations and their results are illustrated.

\subsection{Evaluation Criteria }
\subsubsection{Dice Similarity}
The loss function used during training is the Dice similarity Coefficient also known as S\o
rensen-Dice Index\footnote{From \url{https://en.wikipedia.org/wiki/SrensenDice\textunderscore coefficient}} which is given in Eq.\ref{eq:dice}.
\begin{equation}
2\times\frac{|X \bigcap Y|}{|X| + |Y|}
\label{eq:dice}
\end{equation}
The Dice similarity coefficient is used to measure the spatial overlap of two segmentation masks, $X$ and $Y$. $|X|$ and $|Y|$ refer to the cardinality of the sets $X$ and $Y$. Dice score is a value between $0$ and $1$ which closer value to $1$ show the more similarity between the samples. The idea is to minimize the 1-Dice score computed between two samples. During training, the Dice score between the predicted binary mask and the ground truth mask is computed and this value is subtracted from $1$. The error is then back propagated to be minimized.
\subsubsection{Jaccard Similarity }
The evaluation function for testing the model is the Jaccard Similarity coefficient also known as Intersection over Union. It is given by the Eq.\ref{eq:jack}.
\begin{equation}
\frac{|A \bigcap B|}{|A \bigcup B|}
\label{eq:jack}
\end{equation}
The Jaccard measure is used to compute the similarity of two samples. This is useful to evaluate the accuracy of the model predictions and thus evaluate the model. We compute the Jaccard score between the ground truth masks and the predicted mask.
\subsection{Experimental Set Models}
\subsubsection{Simple U-Net}
In the first set, we implemented a simple U-Net architecture with lesser number of convolution filters than the original U-Net\cite{14} although the architecture was similar in terms of the number of convolution blocks. This model was trained with no data augmentation for $100$ epochs, batch size of $8$ on $384 \times 384$ images. The train-test split was set to $90-10$. The optimizer used was the Adam with learning rate set to $1e-6$. The model was tested on $326$ scans of the test set. The time taken to train was about $60$ to $90$ minutes.

\subsubsection{VGG16-UNet without Data Augmentation} 
In the second set of experiments we implemented the VGG16-UNet. We trained the model for $200$ epochs with no data augmentation. Batch size set was set to $5$ since the network was a bit more dense and had more than $100$ million trainable parameters. The optimizer function used is the SGD(Stochastic Gradient Descent) with the learning set to $0.01$. Here the convolution kernel size is set to $5 \times 5$ and the stride length of $1$ to generate overlapping feature maps. The training set images were resized to $224 \times 224$ due to memory constraints. The train-test split was set to $90-10$. The model was tested on resized $224 \times 224$, $326$ scans of the test set. The time taken to train this model is about $170$ minutes.
\begin{figure}
	\centering
	\subfloat[]
	{
		\includegraphics[width=0.35\textwidth]{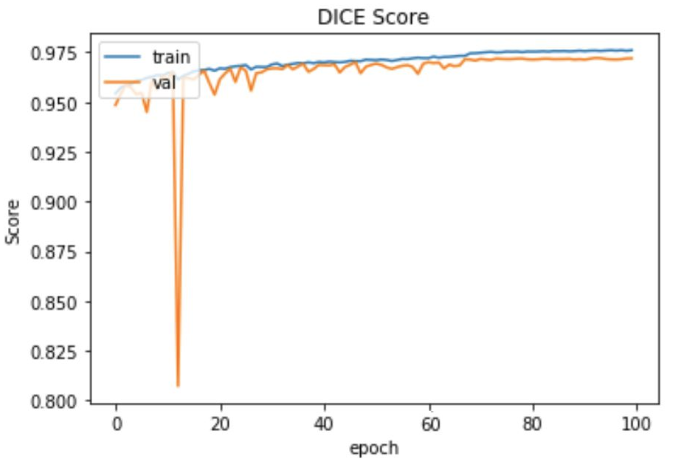}
		\label{fig:p-1}
	}
	\subfloat[]
	{
		\includegraphics[width=0.35\textwidth]{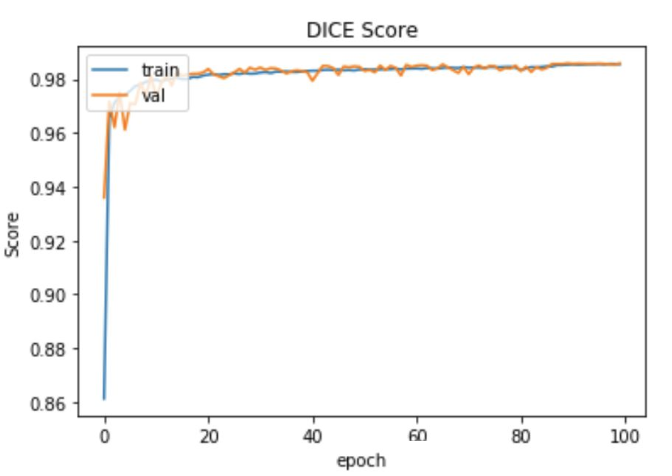}
		\label{fig:p-4}
	}
	\caption{VGG16-UNet model performance plots for lumen and media. (a) Dice score plot for lumen. (b) Dice score plot for media. }
	\label{fig:plots}
\end{figure}
\subsubsection{VGG16-UNet with Data Augmentation} 
In the third set of experiments we continued experimenting with our VGG16-UNet. Here we trained the model with data augmentation. The model was trained for $100$ epochs on $224 \times 224$ images with batch size set to $5$. The same optimizer as before, SGD(Stochastic Gradient Descent) was used and learning rate was set to $0.01$. Here the learning rate was reduced if there was no improvement in validation loss for $20$ epochs. The train-test split was set to $80-20$. Just like before the model was tested on $224 \times 224$, $326$ scans of the test set. This model takes about $170$ minutes to train as well. 
The Dice score and loss plot after training the proposed mode for lumen and media segmentation are presented in Fig.\ref{fig:plots}(a) to (d) respectively. 
\subsection{Results}
Two type of samples for lumen and Media are used for evaluation of the three experimental models that are used in this paper. The first sample has more artifacts and the second one is the regular image due to evaluate the models in different situations. The visual comparison of these three experiments segmentation result are presented in Table .\ref{tbl:visres}. The quantitative results based on the evaluation criteria is presented in Table.\ref{tab:qres}. Furthermore, the general comparison of all of these experiments over the test set IVUS scans are shown in Table.\ref{tab:genqres} which is the average of Jaccard  measure and Dice for all of the test images. The Exp 1, Exp 2 and Exp 3 are referring to simple U-Net,VGG16-UNet without data augmentation and VGG16-UNet with data augmentation. 

\subsubsection{Simple U-Net}
For the first experiment, we can see in Table.\ref{tbl:visres}, that the model does not seem to provide good segmentation results for the noisy images. It is not able to generalize well and the effect of artifacts in the scan like shadowing greatly disturbs the segmentation result. However, the model is able to get good predictions for the regular images. It is able to classify the pixels well since there is not much distortions in the image. 
On an average the results in Table.\ref{tab:qres}, this model is not very good for lumen segmentation. The best Jaccard Similarity score we get for media which is $0.86$. The average Jaccard and Dice score of this experiment for lumen and media segmentation is in Table.\ref{tab:genqres} which is not good.

\subsubsection{VGG16-UNet without Data Augmentation}
From Table.\ref{tbl:visres} we can see that the network is able to generalize quite well. But it still is not able to handle artifacts (in noisy lumen and noisy media). This is probably because the model still tries to classify some black pixels in the vicinity to be as part of the foreground. Some post processing step like thresholding can be used to eliminate any outliers that surround the region of interest. This model is better than the simple U-Net described earlier and provides better segmentation result which is shown in Table.\ref{tab:qres}. Here, the best segmentation result is for media which has jaccard measure of $0.91$. However it still does not perform well enough because the network needs more data. Moreover it is still susceptible to serious artifacts. The average Jaccard and Dice score of this experiment for lumen and media segmentation in  Table.\ref{tab:genqres} is higher than the simple U-Net. 
\begin{table}
	\caption{Visual segmentation Results for different experiments on two sample images for Lumen and Media (one regular and one noisy sample).}
	\label{tbl:visres}
	\centering
	\begin{tabular}{cM{15mm}M{15mm}M{15mm}M{15mm}M{15mm}}
		\toprule
		\textbf{Test Images} & \textbf{Original Image} & \textbf{Grand Truth} & \textbf{Simple U-Net} & \textbf{VGG16-UNet without DA}& \textbf{VGG16-UNet with DA} \\
		\midrule
		\textbf{Noisy Lumen Sample} & \includegraphics[width=1\linewidth]{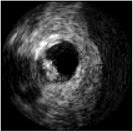} & \includegraphics[width=1\linewidth]{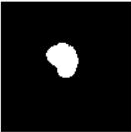} & \includegraphics[width=1\linewidth]{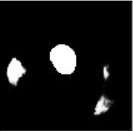} & \includegraphics[width=1\linewidth]{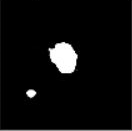}  & \includegraphics[width=1\linewidth]{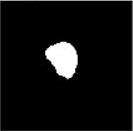} \\
		\textbf{Lumen Sample} & \includegraphics[width=1\linewidth]{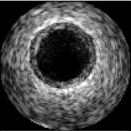} & \includegraphics[width=1\linewidth]{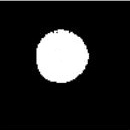} & \includegraphics[width=1\linewidth]{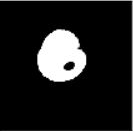} & \includegraphics[width=1\linewidth]{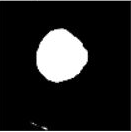}  & \includegraphics[width=1\linewidth]{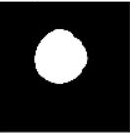} \\
		\textbf{Noisy Media Sample} & \includegraphics[width=1\linewidth]{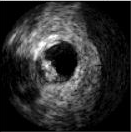} & \includegraphics[width=1\linewidth]{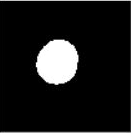} & \includegraphics[width=1\linewidth]{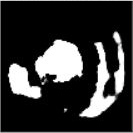} & \includegraphics[width=1\linewidth]{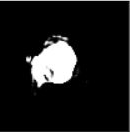}  & \includegraphics[width=1\linewidth]{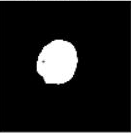} \\
		\textbf{Media Sample} & \includegraphics[width=1\linewidth]{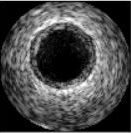} & \includegraphics[width=1\linewidth]{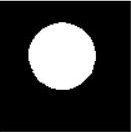} & \includegraphics[width=1\linewidth]{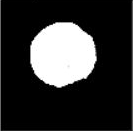} & \includegraphics[width=1\linewidth]{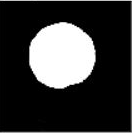}  & \includegraphics[width=1\linewidth]{Pics/nm5.png}  \\
		\bottomrule
	\end{tabular}
\end{table}
 
\begin{table}
	\centering
	\caption{Comparison of different experimental results on two sample images for Lumen and Media (one regular and one noisy sample) based on the quantitative evaluation criteria.}
	\label{tab:qres}
	\begin{adjustbox}{width=1\textwidth}
		\renewcommand{\arraystretch}{1.2}
		\begin{tabular}{|p{5cm}|c|c|c|c|c|c|c|c|c|c|c|c|c|}
			\hline
			\multirow{2}{4cm}{\textbf{Test Images}} & \multicolumn{3}{c|}{\textbf{Intersection Area}} &  \multicolumn{3}{c|}{\textbf{Unit Area}}& \multicolumn{3}{c|}{\textbf{Jaccard}}&  \multicolumn{3}{c|}{\textbf{Dice}}\\
			\cline{2-13}
			& \textbf{Exp 1} & \textbf{Exp 2} & \textbf{Exp 3} & \textbf{Exp 1} & \textbf{Exp 2} & \textbf{Exp 3} & \textbf{Exp 1} & \textbf{Exp 2} & \textbf{Exp 3} & \textbf{Exp 1} & \textbf{Exp 2} & \textbf{Exp 3}\\
			\hline
			\textbf{Noisy Lumen Sample}  & 4539 & 1537 & 2168 & 6448 & 2309 & 2355 & 0.6532 & 0.6656 & 0.9205 & 0.7909 & 0.7992 & 0.9586 \\ \hline
			\textbf{Lumen Sample} & 12773 & 5431 & 5959 & 18995 & 6379 & 6418 & 0.6724 & 0.8513 & 0.9284 & 0.8041 & 0.9191 & 0.9629 \\ \hline
			\textbf{Noisy Media Sample} & 9285 & 2619 & 3534 & 17534 & 4323 & 4299 & 0.5280 & 0.6058 & 0.8220 & 0.6911 & 0.7545 & 0.9023 \\ \hline
			\textbf{Media Sample} & 26314 & 9382 & 9860 & 30532 & 10256 & 10258 & 0.8618 & 0.9147 & 0.9553 & 0.9257 & 0.9554 & 0.9771 \\ \hline
		\end{tabular}
	\end{adjustbox}
	
\end{table}
\begin{table}
	\centering
	\caption{General comparison of all the experiments based on the average quantitative evaluation criteria.}
	\label{tab:genqres}
	\renewcommand{\arraystretch}{1.2}
	\begin{tabular}{|p{2.5cm}|c|c|c|c|c|c|c|c|c|c|c|c|c|}
		\hline
		\multirow{2}{4cm}{\textbf{Test Images}} & \multicolumn{3}{c|}{\textbf{Average Jaccard}} &  \multicolumn{3}{c|}{\textbf{Average Dice}}\\
		\cline{2-7}
		& \textbf{Exp 1} & \textbf{Exp 2} & \textbf{Exp 3} & \textbf{Exp 1} & \textbf{Exp 2} & \textbf{Exp 3}\\
		\hline
		\textbf{Lumen}  & 0.5497 & 0.6965 & 0.7982 & 0.6931 & 0.8129& 0.8846 \\ \hline
		\textbf{Media}  & 0.5754 & 0.7409 & 0.8085 & 0.7197 & 0.8393 & 0.8825\\ \hline
	\end{tabular}
\end{table}

\subsubsection{VGG16-UNet with Data Augmentation}
The last column of Table.\ref{tbl:visres} refers to the proposed model which is shown that it is able to generalize quite well and has learned to eliminate the outliers. Note that at this point we are not performing any preprocessing steps or post processing steps.
The segmentation accuracy has improved significantly as compared to the simple U-Net and VGG16-UNet without data augmentation. In addition, some post processing can be used to improve the boundary smoothness of the region of interest. The Jaccard similarity measures between the ground truth and the prediction is very high based on the presented result in Table.\ref{tab:qres} which is $0.95$. The model has seen a lot of augmented examples is able to generalize very well. Also, Table.\ref{tab:genqres} shows the general scores of the proposed for VGG16-UNet model which has improvement over two previous methods and can get better results for media segmentation than the lumen.

\subsection{Discussion}
We initially experimented with the simple U-Net which the obtained results are not very good. The model performs poorly and is unable to generalize well enough. Besides the amount of data in the training set is quite less and the model does not see many samples to learn. 

The VGG16 U-Net is able to generalize very well and no image from the test set is used to train the model. The results are significantly better the simple U-Net described before. From our literature review the authors\cite{8} who implemented a Hough-CNN to segment MRI and ultrasound modalities of the brain say that "deeper neural networks can work very well with small datasets". We were able to verify this claim by implementing a deep U-Net structure with a VGG16\cite{15} encoder. We applied data augmentation to the training and the results improved significantly since the model has seen a variety of samples. We also conducted an experiment to check whether the Adam\cite{20} or the SGD parameter optimizer gives us better convergence and better accuracy. We found out that SGD works well for this type of problem. Adaptive optimization need not always be a good choice. We used the SGD and we set a high learning rate and reduced it if no improvement was seen after $20$ epochs. This greatly helped the model to converge and improve the accuracy and the Dice score. Our model is able to generate predictions on $326$ test images in the test in about less than $30$ seconds during the inference phase.

Even though our method provides $90\%$ accuracy for some predictions, there are some predictions that are affected by serious artifacts. Moreover the test dataset has more noise in terms of bifurcations, side branches, shadows caused by the catheter. Many images in the test set have more than one class of artifacts. However, Post processing the output images can significantly improve the accuracy.

\section{Conclusion}
This paper have presented a deep learning method for detection of lumen and media in IUVS images which is important to identify the build up plaques in the walls of the coronary vessels. In particular the deeper version of U-Net, a variant of the fully-convolutional neural network that employs a VGG16 as encoder (VGG16-UNet) is used for IUVS image segmentation. The proposed method is evaluated by using the dataset consist of $435$ IVUS scans and each with its own lumen and media labels. We show that the proposed  VGG16-UNet network is able to perform quite well during training and testing and provides good segmentation results visually and quantitatively during the inference phase.

\end{document}